\documentclass[conference]{IEEEtran}
\usepackage{cite}
\usepackage{amsmath, amssymb, amsfonts}
\usepackage{algorithmic}
\usepackage{graphicx}
\usepackage{textcomp}
\usepackage{xcolor}
\usepackage{multirow, multicol}
\def\BibTeX{{\rm B\kern-.05em{\sc i\kern-.025em b}\kern-.08em
    T\kern-.1667em\lower.7ex\hbox{E}\kern-.125emX}}

\begin{document}

\title{Conceptualizing Multi-scale Wavelet Attention and Ray-based Encoding for Human-Object Interaction Detection}

\author{
    \IEEEauthorblockN{Quan Bi Pay\IEEEauthorrefmark{1}, Vishnu Monn Baskaran\IEEEauthorrefmark{1}, Junn Yong Loo\IEEEauthorrefmark{1}, KokSheik Wong\IEEEauthorrefmark{1} and Simon See\IEEEauthorrefmark{2}}
    \IEEEauthorblockA{\IEEEauthorrefmark{1}School of Information Technology, Monash University Malaysia}
    \IEEEauthorblockA{\IEEEauthorrefmark{2}NVIDIA AI Technology Center}
    \IEEEauthorblockA{\{quan.pay, vishnu.monn, loo.junnyong, wong.koksheik\}@monash.edu, ssee@nvidia.com}
}

\maketitle

\begin{abstract}
    Human-object interaction (HOI) detection is essential for accurately localizing and characterizing interactions between humans and objects, providing a comprehensive understanding of complex visual scenes across various domains. However, existing HOI detectors often struggle to deliver reliable predictions efficiently, relying on resource-intensive training methods and inefficient architectures. To address these challenges, we conceptualize a wavelet attention-like backbone and a novel ray-based encoder architecture tailored for HOI detection. Our wavelet backbone addresses the limitations of expressing middle-order interactions by aggregating discriminative features from the low- and high-order interactions extracted from diverse convolutional filters. Concurrently, the ray-based encoder facilitates multi-scale attention by optimizing the focus of the decoder on relevant regions of interest and mitigating computational overhead. As a result of harnessing the attenuated intensity of learnable ray origins, our decoder aligns query embeddings with emphasized regions of interest for accurate predictions. Experimental results on benchmark datasets, including ImageNet and HICO-DET, showcase the potential of our proposed architecture. The code is publicly available at [https://github.com/henry-pay/RayEncoder].
\end{abstract}

\section{Introduction}\label{sec:intro}
In the domain of computer vision, understanding complex visual scenes is a fundamental task. Although popular fields such as object detection~\cite{detr, yolo} identify entities and action recognition~\cite{real-time-action, action-rec} discern specific human actions, they often present a fragmented view of the visual narrative. On the other hand, human-object interaction~\cite{hierarchical-hoi, hier-rcnn} focuses on human-centric interactions with objects, delving deeper into contextual understanding by predicting a set of {\em $<$human, verb, object$>$} triplets within an image~\cite{hoi-set}. This structural understanding is important for high-level semantic understanding tasks such as image captioning~\cite{image-caption} and visual question answering~\cite{vqa} and has a wide range of applications in surveillance systems~\cite{surveillance} and autonomous driving~\cite{autonomous-driving}.\par
Recent progress in object detection has highlighted the convolution-transformer architecture~\cite{detr, deformable-detr}, which offers a simplified structure without relying on hand-crafted components. Object detectors, often used for HOI detection due to similar output representations, typically employ off-the-shelf pre-trained convolutional~\cite{resnet} or transformer-based~\cite{swin} backbones. The aim is to extract highly relevant features before feeding them into transformer-based HOI detectors. The effectiveness of HOI detectors is believed to be closely linked to the robustness of these backbones. However, most HOI detectors are not calibrated to produce reliable and efficient predictions, often requiring costly training measures to achieve ideal detection performance. For instance, DEtection TRansformer (DETR)-based architectures focus on query enhancements~\cite{qahoi, ernet}, such as additional token embeddings, to provide clearer semantic queries for learning contextual information. Such modification often increases computational cost and neglects the efficiency bottleneck within the encoder, thus undermining the efficiency. Similarly, structure-based enhancements that utilize visual-language knowledge~\cite{thid, rlip} or graph neural networks~\cite{spatially-graph, pose-graph} to refine feature extraction often incur computational overhead with additional modules.

\begin{figure*}[tb]
    \centering
    \includegraphics[width=\linewidth]{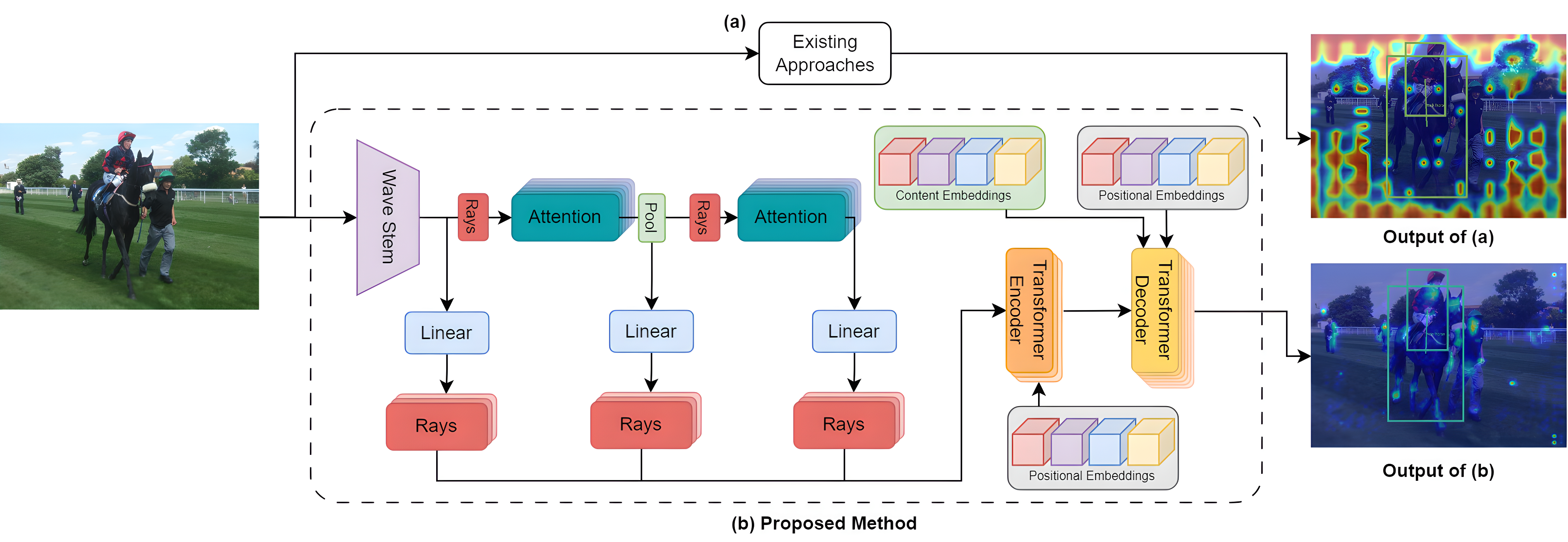}
    \caption{Comparison between (a) existing architecture and (b) proposed architecture in extracting features. The heatmaps are generated using the last layer from the backbone module for a fair comparison. In addition, (b) shows a high-level graphical illustration of the proposed architecture consisting of a wavelet backbone and a ray-based module. The decoder in (b) is adapted from the FGAHOI detector~\cite{fgahoi}.}
    \label{fig:architecture}
\end{figure*}

Apart from inefficiency, the aforementioned techniques often underestimate the ability of convolutional neural networks (CNNs) to extract discriminative features for the representation learning of HOI detectors. Fig.~\ref{fig:architecture}(a) depicts an example of this problem in which current CNN networks~\cite{convnet} draw inspiration from the patch-based approach of vision transformers (ViTs) to perform feature extraction using a hybrid local-global approach. However, a patch-based convolutional approach often encodes extremely low-order or high-order interactions and subsequently ignores middle-order interactions~\cite{bottleneck-dnn}. These middle-order interactions represent the intricate relationships and contextual semantics between humans and objects, which go beyond simple visual features. Hence, they are essential for producing rich feature maps for complex vision tasks such as HOI detection. This shortcoming highlights the need for a redesign of backbone architectures, tailored specifically to the unique challenges of HOI detection. A well-designed backbone must meet key requirements: identifying rich semantic features of humans and objects across scales, and capturing the complex structural and contextual relationships. Meeting these needs can improve HOI detection, enabling more accurate and context-aware systems.

\par
To address the aforementioned challenges, this paper conceptualizes a wavelet attention-like backbone for HOI detection. By leveraging the similarities between multi-resolution analysis and CNN, varying-sized filters are applied and encapsulated with convolutional modulation to extract low-order and high-order interactions, which are then aggregated to obtain middle-order interactions. This technique efficiently consolidates overlooked interactions into discriminative attention maps. In addition, we designed a novel ray-based encoding technique to enhance feature maps to express unique features corresponding to each object. Intuitively, learnable ray origins act as pivotal points to determine regions of interest, emphasizing interactions based on attenuation of ray intensity. This guides the decoder for accurate predictions by aligning query embeddings with regions of interest. Fig.~\ref{fig:architecture}(b) illustrates the proposed architecture, in which, to the best of our knowledge, this is the first utilization of ray-relevant concepts in computing attention within HOI detection.

Our contributions can be summarized as follows:
\begin{itemize}
    \item We first put forward a multi-scale wavelet attention-like backbone that leverages middle-order pixel interactions to extract discriminative features efficiently. The feature extraction and refinements of the proposed wavelet attention mechanism generate rich discriminative features to guide HOI semantic and representation learning.
    \item Next, we propose a novel ray-based encoding mechanism that activates important features to further enhance the prediction of the decoder through high-quality attention maps. The ray's source (or origin) is a learnable parameter to preemptively encode initial attention proposals that are subsequently refined in the decoder for accurate HOI prediction. This technique also reduces the number of encoder layers, thus improving the efficiency of the model.    
    \item Then, we evaluate the proposed backbone architecture on ImageNet with a detailed analysis, showcasing its potential in extracting rich feature maps. We also assess the performance of the proposed architecture on HICO-DET, the largest benchmark dataset for HOI, to demonstrate its potential application in the HOI domain.
\end{itemize}

\section{Related Works}\label{sec:background}
\subsection{HOI Detection}\label{sec:hoi}
In previous works, off-the-shelf object detection models, such as Faster-RCNN~\cite{faster-rcnn}, are employed to identify and localize humans and objects in the image. These entities are paired to form the human-object pairs and sent to subsequent modules for interaction predictions. Given that the first stage offers limited optimization, efforts are focused on refining the prediction of the interactions through additional features or external knowledge. Graph neural networks~\cite{qi2018learning} and multi-stream architectures~\cite{hou2020visual, kim2020uniondet} are the best representative works to utilize additional features in enhancing HOI predictions. On the other hand, visual-language models~\cite{radford2021learning} are mostly leveraged to improve the detection accuracy using external language knowledge. However, the quality of proposals in the first stage is not guaranteed under most scenarios. Hence, such methods cannot achieve an optimal solution. More importantly, two-stage approaches are limited in speed due to their serial architecture. This led to the development of one-stage methods using strong feature representation learning to perform parallel human-object pair detection and interaction prediction. The advent of DETR~\cite{detr} popularized the use of modern one-stage approaches~\cite{hoi-set, ernet, qahoi} in HOI detection as it allows end-to-end training without hand-crafted components. These approaches employ a convolution-transformer architecture that dynamically focuses on the most relevant sets of human-object features, resulting in more accurate HOI predictions~\cite{hoi-set}. Recent developments in DETR-based architecture can be categorized into two classes: (1) query-enhancement~\cite{qahoi, fgahoi, ernet}, which aims at introducing additional query tokens or mechanisms that enhance the quality of query tokens, and (2) structure-based enhancement~\cite{thid, rlip}, which aims at customizing model architectures for HOI detection. In addition, concepts and techniques from visual-language models~\cite{radford2021learning} are leveraged in structure-based customization to learn the expressive HOI instance representation~\cite{zhou2024dual}. Meanwhile, high-quality query tokens can resolve noisy backgrounds~\cite{fgahoi} and long-tail distribution~\cite{ernet} issues in HOI detection. Although one-stage approaches are generally more efficient than two-stage approaches, their application in real-world scenarios is still limited due to the encoder's efficiency bottleneck in encoding multi-scale interactions.\par
\subsection{Wavelet-based Architecture}\label{sec:wavelet}
Wavelets were first introduced in multi-resolution analysis, revolutionizing approaches to data compression and texture discrimination by providing a framework for representing information at different scales and resolutions. Wavelets localize both in frequency and spatial domains, making them particularly suitable for analyzing signals and images with spatially varying features. This property offers a higher degree of interpretability, which is especially valuable in application domains where reliability, transparency, and explainability are critical considerations~\cite{dawn}. Wavelet-based deep learning architectures like WCNN~\cite{fujieda2018wavelet} and DAWN~\cite{dawn} exploited the intrinsic similarity between multi-resolution analysis and convolution operations to initiate learning directly within the wavelet domain. This fusion marked a significant step forward, leading to wavelet-based architectures being successfully applied in a variety of tasks, including image compression~\cite{Akyazi_2019_CVPR_Workshops}, image super-resolution~\cite{Huang_2017_ICCV}, image denoising~\cite{winnet}, and image classification~\cite{dawn}. Recently,~\cite{yang2023explainable} proposed to combine attention with wavelets. High-frequency and low-frequency components in wavelets were utilized to model and capture global structural information, while the attention mechanism was employed to dynamically focus on the most informative features, enhancing the model's interpretability and performance across various tasks. To this end, multi-resolution analysis is inherently well-suited for extracting multi-scale feature maps while avoiding information redundancy, which represents the primary bottleneck in the multi-scale deformable attention mechanism.\par
\subsection{Ray-based Architecture}\label{sec:ray}
Ray-based deep learning architectures are widely used for neural scene representations and neural rendering, offering solutions to the inherent limitations of traditional and voxel-based models~\cite{ sitzmann2021light}. Traditional models often struggle with scalability and memory efficiency, while voxel-based approaches face challenges due to their high computational cost and fixed resolution. In contrast, ray-based architectures enable flexible and efficient scene encoding by operating directly in a continuous space, leveraging differentiable rendering to effectively reconstruct scenes from sparse and incomplete data. The Light Field Network~\cite{sitzmann2021light} has emerged as a notable advancement in this domain, utilizing 360-degree light fields to represent scenes comprehensively. This method reduces the dependency on extensive datasets by capturing multi-view scene geometry and appearance in a compact form. Moreover, recent work, such as Pointersect~\cite{chang2023pointersect}, extends these principles by directly leveraging 3D spatial information from point clouds. By bypassing the need for per-scene optimization, Pointersect achieves efficient scene rendering while preserving fine-grained structural details, further broadening the applicability of ray-based architectures for real-time and scalable rendering tasks. These methods were not applied in 2D image analysis due to lack of depth information, but incorporating attention mechanisms opens new possibilities for 2D ray-based architectures, as discussed in subsequent sections.
\section{Methodology}\label{sec:methodology}
\subsection{Wavelet Backbone}\label{sec:wavelet-backbone}
\subsubsection{Background}
Given an input image, $x\in\mathbb{R}^{C\times H\times W}$,  the convolution $(*)$ and pooling $(\downarrow)$ operations within CNN~\cite{cnn} are represented as 
\begin{equation}
    \label{eq:cnn}
    \hat{y} = \left(x*k\right)\downarrow p,
\end{equation}
where $k\in\mathbb{R}^{C_{\text{out}}\times C\times d\times d}$ represents the filter window and $p\in\mathbb{N}^+$ determines the order of pooling. Meanwhile, multi-resolution analysis utilizes~\eqref{eq:cnn} with a pair of quadrature mirror filters, namely low-pass and high-pass filters, to repeatedly decompose $x$ into low- and high-frequency information.

A high-pass filter captures the fine details in an image, such as edges, and encodes simple color and pattern distributions. In contrast, a low-pass filter emphasizes larger features and background elements, giving a broader view of the image. CNNs leverage these characteristics with filters of different sizes, allowing them to capture a range of information from the fine-grained to the global. This observation suggests that a straightforward way to model intermediate interactions within CNN layers is to use filters of varying sizes.

\subsubsection{Architecture}
\begin{figure*}[tb]
    \centering
    \includegraphics[width=\linewidth]{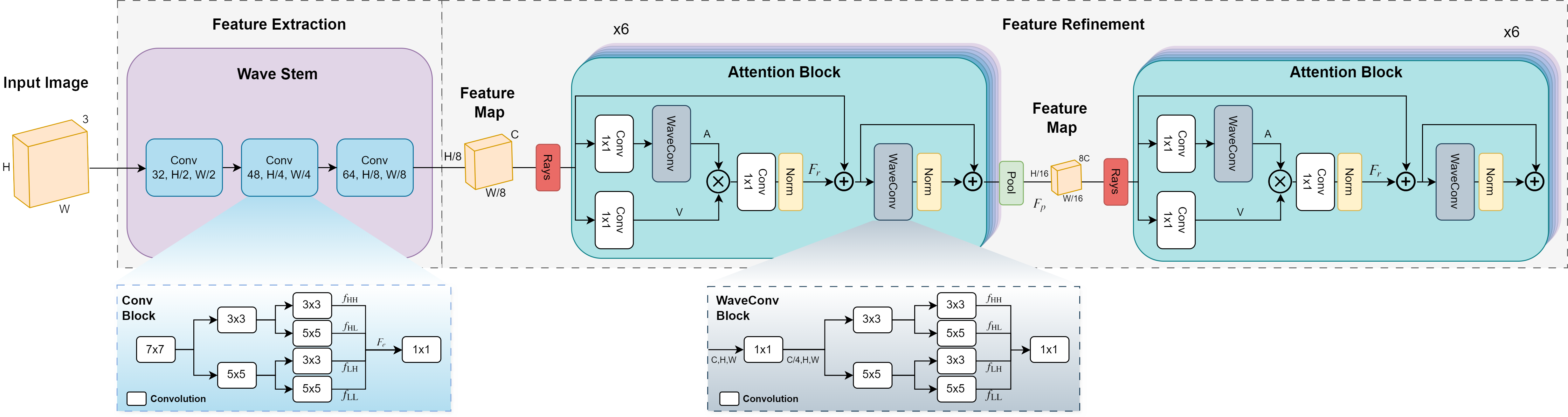}
    \caption{Wavelet backbone architecture is divided into 2 stages, feature extraction and feature refinement, to obtain focused, rich semantic features. Note that the feature refinement stage is repeated twice, with each stage containing 6 blocks of convolutions before the pooling module.}
    \label{fig:backbone}
\end{figure*}

The backbone architecture, as illustrated in Fig.~\ref{fig:backbone}, adopts a pyramidal structure to extract multi-scale features. It is crucial to recognize features of varying sizes and details to ensure an effective information flow from lower to higher layers, where deeper layers become responsive to larger regions of the image. Consequently, this design facilitates the integration of contextual information from a broader receptive field, enriching semantic features.

{\bf Feature extraction} retrieves significant features in a data-compression fashion. This facilitates a fast spatial reduction to save memory and computational costs in subsequent layers. To achieve this, we use a $7\times 7$ convolutional filter to capture features in a wider field of view. This provides an initial understanding of the overall image structure, which is crucial for spatial compression in later stages. Subsequently, multi-resolution analysis is leveraged to generate feature maps at different interaction orders, expressed as
\begin{equation}
    \label{eq:wavelet}
    F_e = \left[f_{\text{LL}};f_{\text{LH}};f_{\text{HL}};f_{\text{HH}}\right].
\end{equation}

A feature map, $f\in\mathbb{R}^{C\times H\times W}$, is processed using the small and large filters to isolate high-frequency and low-frequency information, denoted as $f_{\text{H}}$ and $f_{\text{L}}$. The resultant feature maps undergo the same treatment to produce 4 maps whose concatenation yields the final feature map, $F_e\in\mathbb{R}^{C_{\text{out}}\times \frac{H}{2}\times \frac{W}{2}}$. This process resembles that of separable filtering, in which each combination isolates a distinct frequency band in multiresolution analysis. With 4 feature maps, we essentially incorporate directional information into each of them to enrich the extracted features. The spatial reduction is realized with a stride of 2 at each decomposition level, either horizontally or vertically. Finally, a point convolution processes linear combinations of channels in a spatially coherent context, enhancing the model's capacity to differentiate nuanced features and patterns. This design prioritizes high-level abstractions over pixel-level details, incorporating essential elements such as edges and blurred representations of large objects, aiming to retain maximal information during spatial compression.

{\bf Feature refinement} aggregates contextual information and modulates the overlooked pixel interactions in feature maps. The modulation is typically computed as a weighted average of the value based on a similarity score $A$, which measures the relationship between each pair of image patches. Consequently, the softmax distribution is utilized to compute the matrix $A$ as in the attention mechanism. In this work, convolutional features are directly used to modulate the feature maps $V$ through element-wise multiplication following~\cite{hou2022conv2former}. The operation can be expressed as
\begin{equation}\label{eq:modulation}
    F_r = \text{Conv}_{1\times 1}\left(A\circ V\right).
\end{equation}

This refines the features based on local neighbourhood correlations to congregate middle-order interactions. To achieve this, a wave convolution layer with bottleneck design~\cite{resnet} is applied to compute $A$. We utilize a point convolution to reduce the channel dimensions by a factor of 4. Following~\eqref{eq:wavelet}, 4 feature maps are generated which can be concatenated to obtain $A$. Note that a convolution operation is a weighted sum of each local patch with the kernel, which can be viewed as an unnormalized dot-product between the patch and the kernel. In this context, the kernel is functioning as a key sliding on the query feature map, computing the relationship between image patches and the known pattern in the kernel. Stacking 2 convolution layers helps expand the receptive fields and efficiently aggregates correlations within the local neighbourhood.

In terms of implementation, depth-wise convolution with a stride of 1, unlike in the feature extraction stage, is employed for efficient computation and retaining spatial information. To avoid over-parametrization, the same filters are reused in the 2-stage decomposition. At the end of the refinement stage, we apply a pooling layer to summarize the fine‑grained local features and eliminate trivial redundancies in the attention maps.  Following Eq.~\eqref{eq:wavelet}, it does so by fusing features from differently sized filters in pairs as $F_p=\left[f_{\text{LL}}+f_{\text{HH}};f_{\text{LH}}+f_{\text{HL}}\right]$. This is passed to the next level for further refinement to detect small-scale objects.\par
\subsection{Ray-based Encoder}\label{sec:ray-encoder}
\subsubsection{Background}
Given a ray, $r\in\mathbb{R}^3$, defined by its origin, $o$ and normalized direction, $d$ as $r = o + td$ where $t$ is a scaling factor, ray tracing algorithms utilize the rendering equation to simulate the path taken by ray as it interacts with the 3D object surfaces. Ray marching algorithms leverage the signed distance function to incrementally approximate their path. However, these methods were not explored in 2D image processing due to the absence of depth information. Hence, classical 2D image processing focuses on pixel value and pattern analysis using kernelized functions. To this end, we propose a method of linking ray-based and attention mechanism~\cite{vaswani2017attention}, which subsequently leads us to ray-based modules for feature encoding.

In a 2D image, $I \in \mathbb{R}^{H \times W}$, we establish a Cartesian coordinate system with the image center positioned at the origin, $(0, 0)$. This arrangement assigns each pixel a coordinate point, $(x, y)$. When a ray source is placed at any point within this system, denoted as $(x_r, y_r)$, the pixels close to the ray source appear brighter, while the brightness diminishes as the distance from the source increases. This resembles the attention mechanism in~\cite{vaswani2017attention}, which emphasizes a part of an image by assigning a weighted score to image patches that would indicate relevance for the final prediction, expressed as
\begin{equation}
    \label{eq:attn}
    A = \text{softmax}\left(\frac{KQ}{\sqrt{d}}\right).
\end{equation}
Therefore, we can integrate a ray-based mechanism in 2D image processing by modelling $K$ and $Q$ based on the ray origin. This design, which depends on the source rather than the ray's trajectory, is not influenced by its direction and does not require depth information. Unlike the attention mechanism based on the learned relationship between the features, the ray source provides prior guidance, indicating the prior distribution of the critical features. This allows the model to leverage inherent spatial or contextual priors from the inputs, enhancing its ability to focus on relevant regions or aspects of the data, ultimately improving performance in tasks requiring precise localization or feature extraction.

\subsubsection{Architecture}
\begin{figure}[tb]
    \centering
    \includegraphics[width=\linewidth]{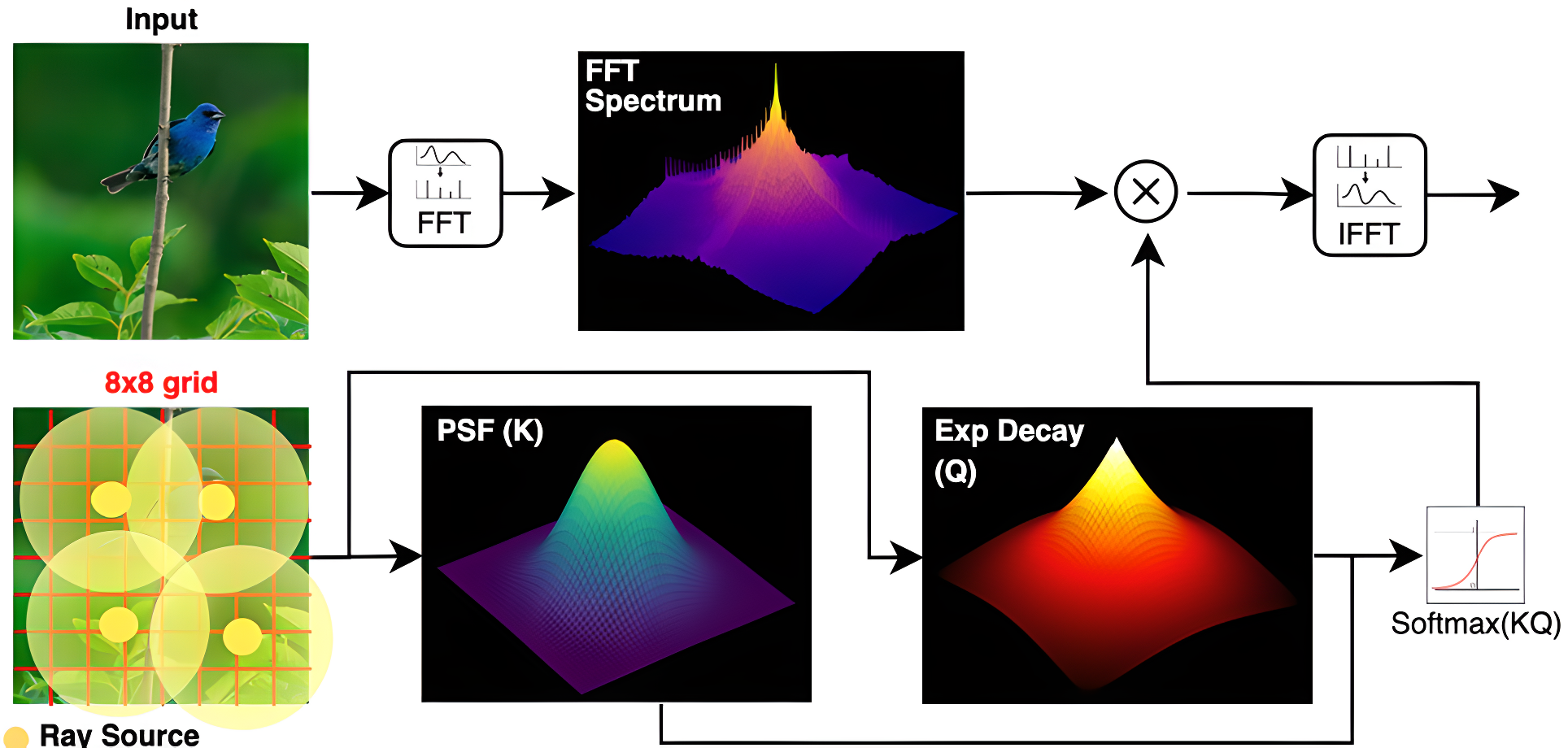}
    \caption{Proposed ray mechanism. 12 ray origins are utilized to produce a combined attenuation map. The resultant attenuation map provides an attention score for each pixel (i, j).}
    \label{fig:ray_architecture}
\end{figure}

Fig.~\ref{fig:ray_architecture} illustrates the proposed ray architecture, which consists of ray-attention and representation learning using 2D Fast Fourier transformation.

{\bf Ray origins} are initialized evenly on a unit circle as learnable parameters, $O\in\mathbb{R}^{n\times 2}$, where $n$ is the number of origin points. This design choice follows the observation that most of the objects are centered in the image as shown in Fig.~\ref{fig:ray_convergence}(d). Note that these origins act as pivotal points for a region of interest where pixels at closer proximity are assigned higher weightage. Consequently, we utilize pairwise distance between the ray origins and pixel coordinates, $C\in\mathbb{R}^{HW\times 2}$, to generate a normalized distance matrix, $D\in\mathbb{R}^{n\times HW}$, expressed as
\begin{equation}
    \label{eq:distance}
    D_{(i, j)} = \|O_i - C_j\|.
\end{equation}

{\bf Point spread function (PSF)} characterizes the intensity distribution of light around its central peak. For simplicity, we model PSF as 2D normalized Gaussian with learnable variance. This spread distribution is treated as key, $K$, as it encodes additional information on the resolution of the ray source, emphasizing the coverage of the ray whereby pixels within the boundary are important.

{\bf Exponential decay} is widely used to model attenuation of light intensity traveling through a medium. In this context, we model how a ray's intensity diminishes as it is absorbed by the pixels it passes through. This produces the query, $Q$, which corresponds to the remaining light intensity that is potentially absorbed by the pixel at position $(i, j)$. Following that, $A'$ forms a ray attenuation map based on the ray origin, expressed as
\begin{equation}
    \label{eq:k}
    A' = \text{softmax}\left(\text{PSF}(D)\cdot \beta\exp\left(-\alpha D\right)\right).
\end{equation}

{\bf Learning in frequency domain} is motivated by the equivalence between element-wise multiplication and depthwise global circular convolution in the spatial domain~\cite{rao2021global}. In particular, pixel interactions are modelled in the frequency spectrum, which is modulated through the ray's attenuation map, $A'$. Theoretically, this approach captures long-term and short-term interactions since the ray origins can cover all frequencies by adjusting the learnable parameters. This method also reduces the prevalent inductive bias in a standard convolution.
\section{Experiments}
\subsection{Overview}\label{sec:experiment-overview}
\subsubsection{Datasets and Evaluation} A series of experiments were carried out to evaluate the performance of the proposed architectures. These experiments were conducted using two datasets: ImageNet for image classification and HICO-DET for HOI detection. We employed accuracy and F1-score as evaluation metrics for image classification, while mean average precision (mAP) was used for HOI detection.

\subsubsection{Implementation Details}
We utilized an AdamW optimizer for 300 epochs of training with a one-cycle cosine decay learning rate scheduler. A batch size of 1,024 and an initial learning rate of 1e-3 with a weight decay of 0.05 were set. Image sizes were standardized for image classification to $224\times 224$, and we followed the same input image scale augmentation in~\cite{touvron2021training}. We utilized the FGAHOI decoder and its training settings~\cite{fgahoi} for HOI detection. As for the encoder, we utilized 3 ray-based encoding layers and 3 transformer encoder layers. We first pretrained the wavelet backbone on ImageNet and then on HICO-DET. All model training and inferences were carried out using dual NVIDIA RTX 8000 GPUs. Table~\ref{tab:implementation} details the model configuration.

\begin{table}[tb]
    \centering
    \caption{Architectural Configuration}
    \label{tab:implementation}
    \begin{tabular}{c|c|c|c}
        \hline
        Stage & Shape & Layer Setting & Wavelet+3 Rays \\
        \hline
        \multirow{3}{*}{Feature Extraction} & \multirow{3}{*}{$\frac{H\times W}{8}$} & Channels & $32\to 48 \to 64$ \\
        \cline{3-4}
        & & Kernel (Low) & 3 \\
        \cline{3-4}
        & & Kernel (High) & 5 \\
        \hline
        \multirow{6}{*}{Feature Refinement} & \multirow{6}{*}{$\frac{H\times W}{16}$} & Channels & $64 \to 512 \to 4096$ \\
        \cline{3-4}
        & & Kernel (Low) & 3 \\
        \cline{3-4}
        & & Kernel(High) & 5 \\
        \cline{3-4}
        & & \# Origins & 12 \\
        \cline{3-4}
        & & \# Ray Layers & $1 + 1$ \\
        \cline{3-4}
        & & \# Blocks & $6 + 6$ \\
        \hline
        \multirow{3}{*}{Ray Encoder} & \multirow{3}{*}{$\frac{H\times W}{16}$} & Channel & $4096 \to 256$ \\
        \cline{3-4}
        & & \# Origins & 12 \\
        \cline{3-4}
        & & \# Layers & 3 \\
        \hline
        DETR Decoder & \multicolumn{3}{c}{Refer to~\cite{fgahoi}} \\
        \hline
    \end{tabular}
\end{table}\par
\subsection{Image Classification}\label{sec:imagenet}
\subsubsection{Quantitative Analysis} 
Table~\ref{tab:imagenet} tabulates the performance of the backbone model on the ImageNet-1K dataset. The proposed multi-scale wavelet backbone achieves a competitive result in image classification, i.e., $73.36\%$ of top-1 accuracy. More importantly, the top-1 accuracy increases to $74.54\%$ when ray encoding is incorporated with the wavelet backbone (i.e., Wavelet+3 Rays). We also observe a similar F1-score where the proposed model performs at a consistent level when predicting each instance, weighted by the class size. 

\begin{table}[t]
    \centering
    \caption{Performance comparison of the proposed backbone model variants on ImageNet Test Set}
    \label{tab:imagenet}
    \begin{tabular}{|c|c|c|c|c|c|c|}
        \hline
        \multirow{2}{*}{Model} & \multirow{2}{*}{params (M)} & 
        \multirow{2}{*}{FPS} & \multicolumn{2}{c|}{Accuracy (\%)} & \multirow{2}{*}{F1-score} \\
        \cline{4-5}
         & & & (top-1) & (top-5) & \\
         \hline
         Wavelet & \textbf{9.58} &
         \textbf{1388} & 73.36 & 91.28 & 0.73 \\
         \hline
         Wavelet+1 Ray & 9.98 &
         895 & 73.71 & 91.43 & 0.73 \\
         \hline
         Wavelet+2 Rays & 9.98 &
         714 & 74.02 & 91.67 & 0.74 \\
         \hline
         Wavelet+3 Rays & 10.38 &
         709 & \textbf{74.54} & \textbf{92.22} & \textbf{0.74} \\
         \hline
    \end{tabular}
\end{table}

\subsubsection{Qualitative Analysis}
\begin{figure}[tb]
    \centering
    \includegraphics[width=\linewidth]{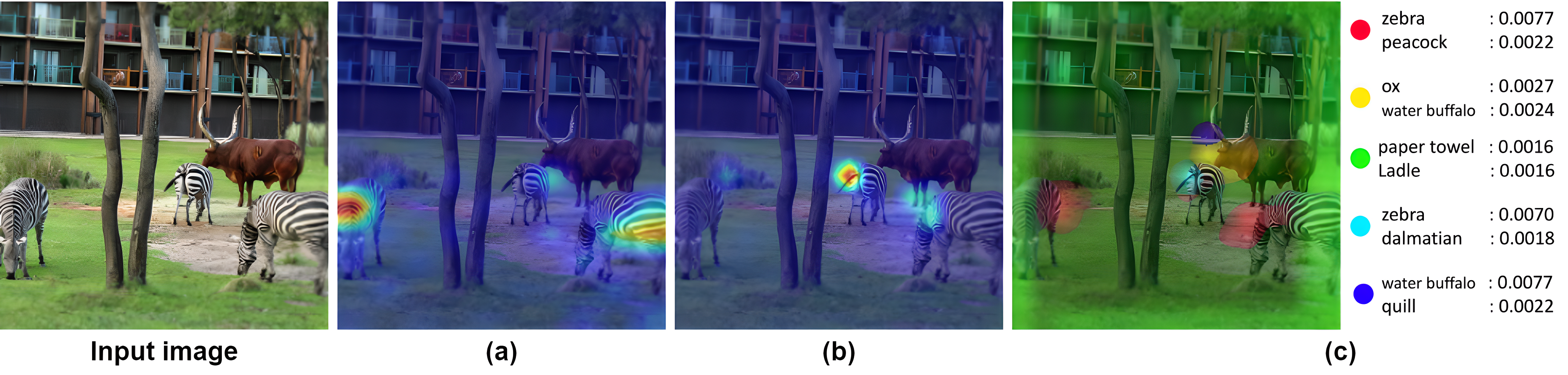}
    \caption{Grad-CAM activation maps on (a) Wavelet and (b) Wavelet+3 Rays. (c) represents a deep feature factorization using five semantic feature clusters, each presented with top-2 classes and their scores.}
    \label{fig:qualitative_imagenet}
\end{figure}

\begin{figure}[tb]
    \centering
    \includegraphics[width=\linewidth]{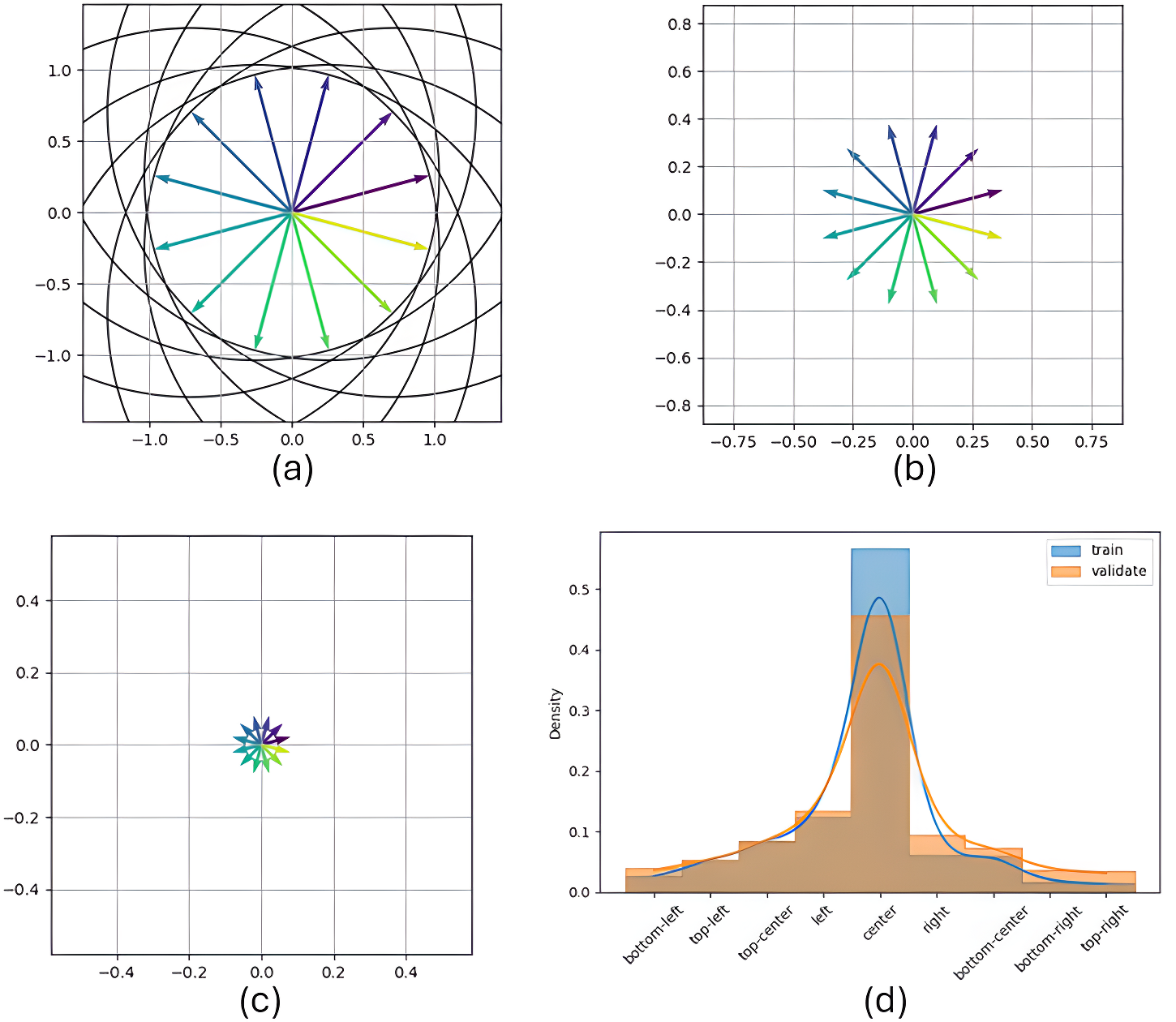}
    \caption{Position of ray sources on a 2D Cartesian plane at (a) epoch 0, (b) epoch 25, and (c) epoch 50. (d) shows the distribution of positions of objects in the ImageNet training set and validation set. Note that most objects in ImageNet appear in the middle. Therefore, the ray sources converge to the center of these images during training.}
    \label{fig:ray_convergence}
\end{figure}

\begin{figure}[tb]
    \centering
    \includegraphics[width=\linewidth]{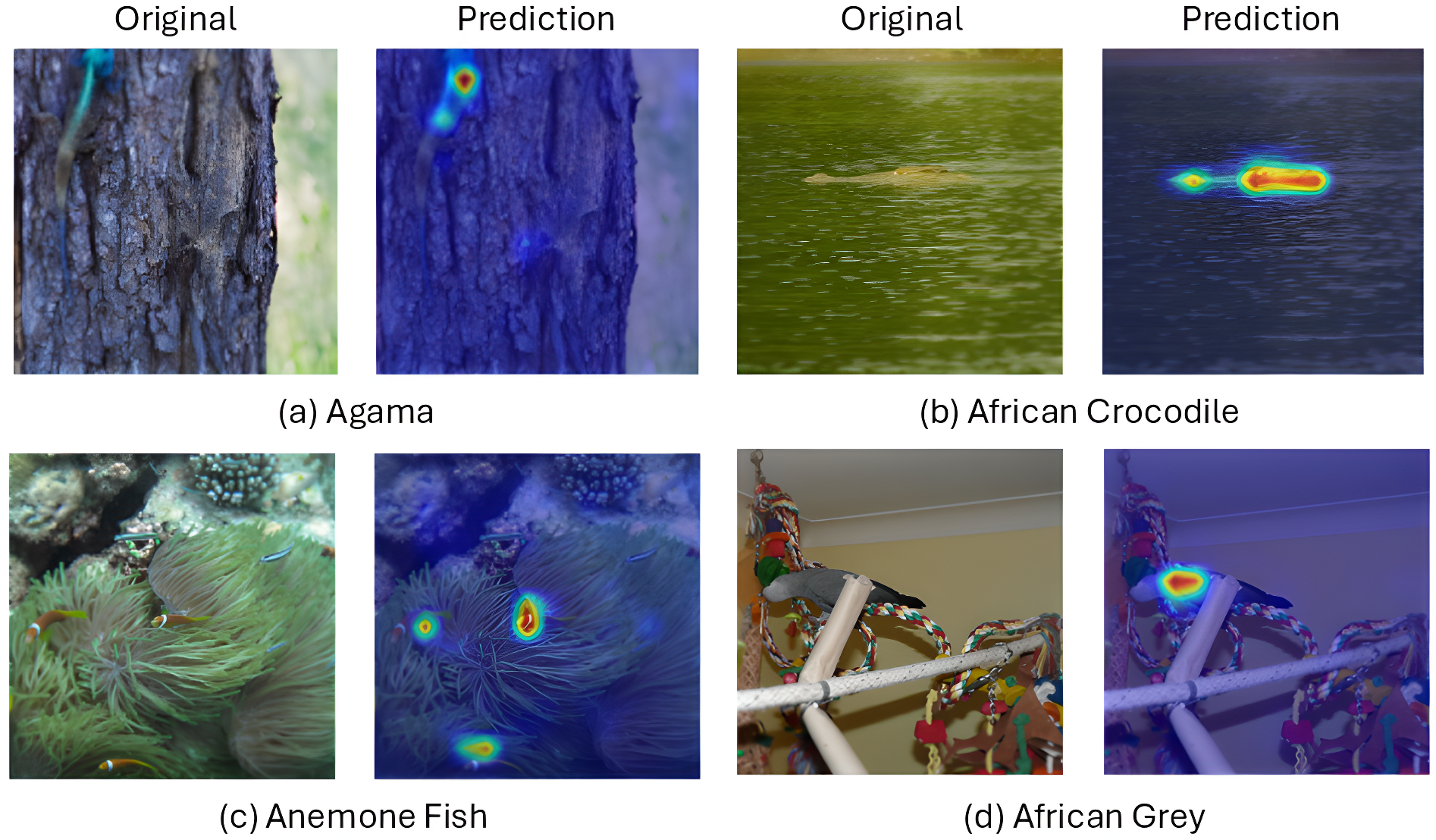}
    \caption{Visualization of Grad-CAM activation maps on different test images from the ImageNet dataset, demonstrating the capability of the proposed backbone in detecting small-scale objects amidst numerous distractions.}
    \label{fig:qualitative_imagenet_extra}
\end{figure}

To better comprehend the advantages of the proposed architecture, we visualize the activation maps generated by Wavelet and Wavelet+3 Rays using Grad-CAM~\cite{selvaraju2017grad}. As seen in Fig.~\ref{fig:qualitative_imagenet}(a), Wavelet dilutes the focus to all of the zebras. Conversely, in Fig.~\ref{fig:qualitative_imagenet}(b), Wavelet+3 Rays densely coagulate around the semantic features, sharpening the model's ability to concentrate on decisive features. However, as depicted in Fig.~\ref{fig:ray_convergence}(a), (b), and (c), the ray-based mechanism shifts the focus of the model from the sides to the middle part of the image. This pattern is expected based on analysis in Fig.~\ref{fig:ray_convergence}(d), where key features of images are concentrated in the center. In this context, a ray-based mechanism provides enhanced emphasis on key features of an image but is susceptible to the context of training data. In spite of this, additional analysis using deep feature factorization~\cite{collins2018deep} as illustrated in Fig.~\ref{fig:qualitative_imagenet}(c) reveals that the ray-based mechanism does not ignore the semantic features of zebras at the sides of the image. In fact, they are the highest-scoring class within respective clusters. This highlights the potential application of a ray-based mechanism in 2D image classification. To further assess the ability of the proposed wavelet backbone in extracting semantic context, selected images with small-scale objects amidst numerous distractions are fed into the model. As shown in Fig.~\ref{fig:qualitative_imagenet_extra}, the proposed model can accurately identify the semantic structure of the predicted object, even when the color of the small-scale object closely resembles and blends with the background elements. Overall, analysis of the activation map shows that a ray-based mechanism aids the backbone by refining feature maps of an image for improved representation learning based on its symmetrical properties.

\subsubsection{Ablation Study}
To assess the effectiveness of ray-based encoding in improving the performance of the wavelet backbone, we ablate the number of ray layers from 0 to 3, as shown in Table~\ref{tab:imagenet}. It has been shown that the addition of ray layers would help to improve the top-1 accuracy of the wavelet backbone. However, as more ray layers are added, the efficiency of the model is affected, indicated by the decreasing FPS. This trade-off highlights the need for a balanced design, where the number of ray layers is optimized to achieve improved accuracy without significantly compromising the model's real-time performance.\par
\subsection{HOI Detection}\label{sec:hico-det}
\subsubsection{Quantitative Analysis} Table~\ref{tab:hico-det} compares the performance of our proposed architecture (ablated on ray modules) with the FGAHOI model on the HICO-DET dataset, which serves as a benchmark. The wavelet backbone achieved a mAP of 20.97\% under default {\it Full} settings. However, with ray encoding layers added, the mAP increases to 24.07\%. Despite not reaching the same mAP as FGAHOI, the proposed model has significantly fewer parameters and higher inferencing speed (i.e., FPS), indicating a more efficient approach to HOI modeling.

\begin{table}[tb]
    \centering
    \caption{Comparison of models validated on the HICO-DET test-set with a fixed input image size of $800 \times 1333$ pixels.}
    \label{tab:hico-det}
    \scalebox{0.72}{\begin{tabular}{|c|c|c|c|c|c|c|c|c|}
        \hline
        \multirow{2}{*}{Model} & \multirow{2}{*}{params (M)} & \multirow{2}{*}{FPS} & \multicolumn{3}{c|}{Default (mAP)} & \multicolumn{3}{c|}{Known Object (mAP)} \\
        \cline{4-9}
        & & & Full & Rare & Non-rare & Full & Rare & Non-rare \\
        \hline
        Wavelet & 36.94 & 9 & 20.97 & 12.41 & 23.53 & 23.69 & 14.47 & 26.45 \\
        Wavelet+3 Rays & \textbf{34.46} & \textbf{10} & 24.07 & 15.94 & 26.37 & 26.42 & 18.36 & 27.58 \\
        \hline
        FGAHOI & 55.42 & 8 & \textbf{29.81} & 22.17 & 32.09 & 32.37 & 24.32 & 34.78 \\
        \hline
    \end{tabular}}
\end{table}

\subsubsection{Qualitative Analysis} 
\begin{figure}[t]
    \centering
    \includegraphics[width=\linewidth]{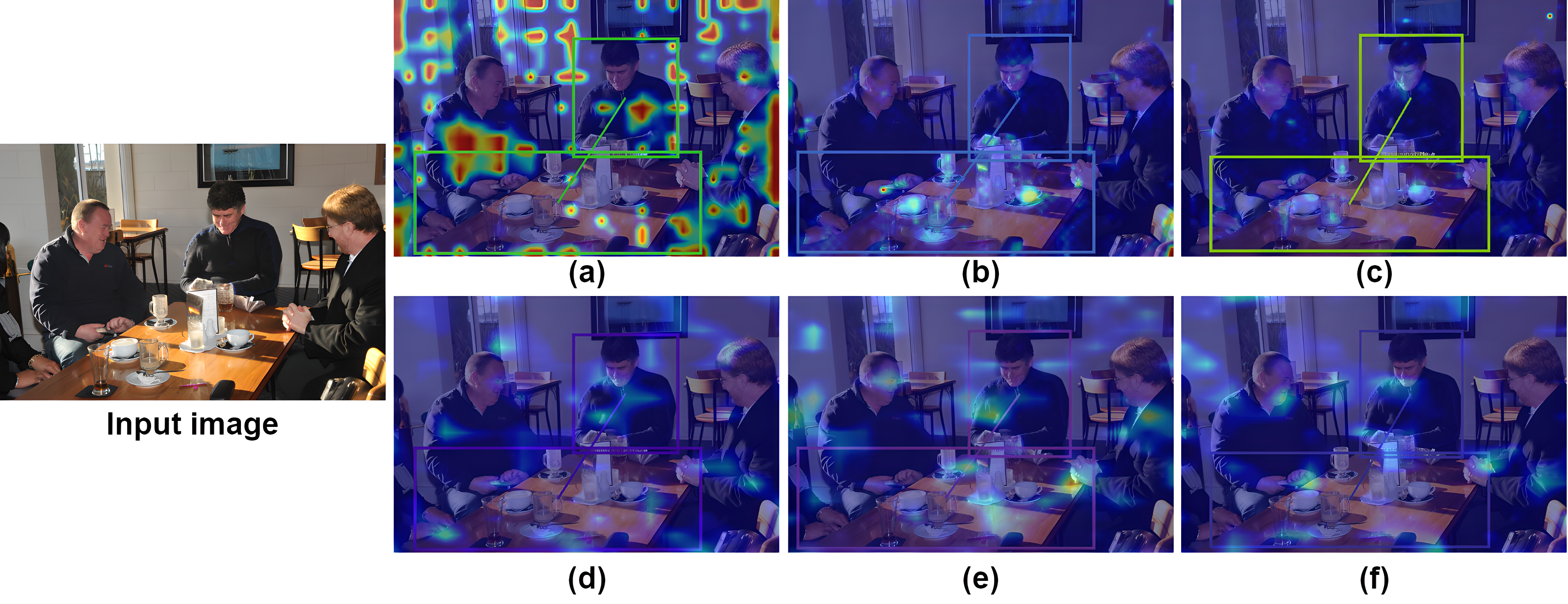}
    \caption{Eigen-CAM~\cite{muhammad2020eigen} activation maps on (a) Swin, (b) wavelet, and (c) wavelet+3 Rays show that wavelet backbone can better identify semantic features. Activation maps on decoder layer (d), (e), and (f) show that semantic features relevant to HOI triplets are more focused on the proposed method.}
    \label{fig:qualitative_hico}
\end{figure}

In Fig.~\ref{fig:qualitative_hico}(a), we observe that the Swin backbone~\cite{swin} used in FGAHOI exhibits scattered focus across the image, with nearly half of the attention on background features. Conversely, the proposed wavelet backbone aggregates middle-order interactions, and as seen in Fig.~\ref{fig:qualitative_hico}(b), this model highlights the utensils on the table. Integrating ray-based encoding with Wavelet backbone shifts part of the attention towards the human, as observed in Fig.~\ref{fig:qualitative_hico}(c), which is advantageous for predicting HOI triplets. Traditionally, HOI detectors rely on the backbone to provide rich features that are refined by the transformer encoder to extract relevant semantics. The decoder would then remove redundancies for HOI detection, as shown in Fig.~\ref{fig:qualitative_hico}(d). Our approach of using a wavelet and ray-based mechanism extracts more focused semantic features. The encoder processes 2D feature maps into sequential data, while the decoder identifies HOI triplets by turning these concentrated semantic features into specific areas for detection, as depicted in Fig.~\ref{fig:qualitative_hico}(e) and (f).

\begin{figure}[tb]
    \centering
    \includegraphics[width=\linewidth]{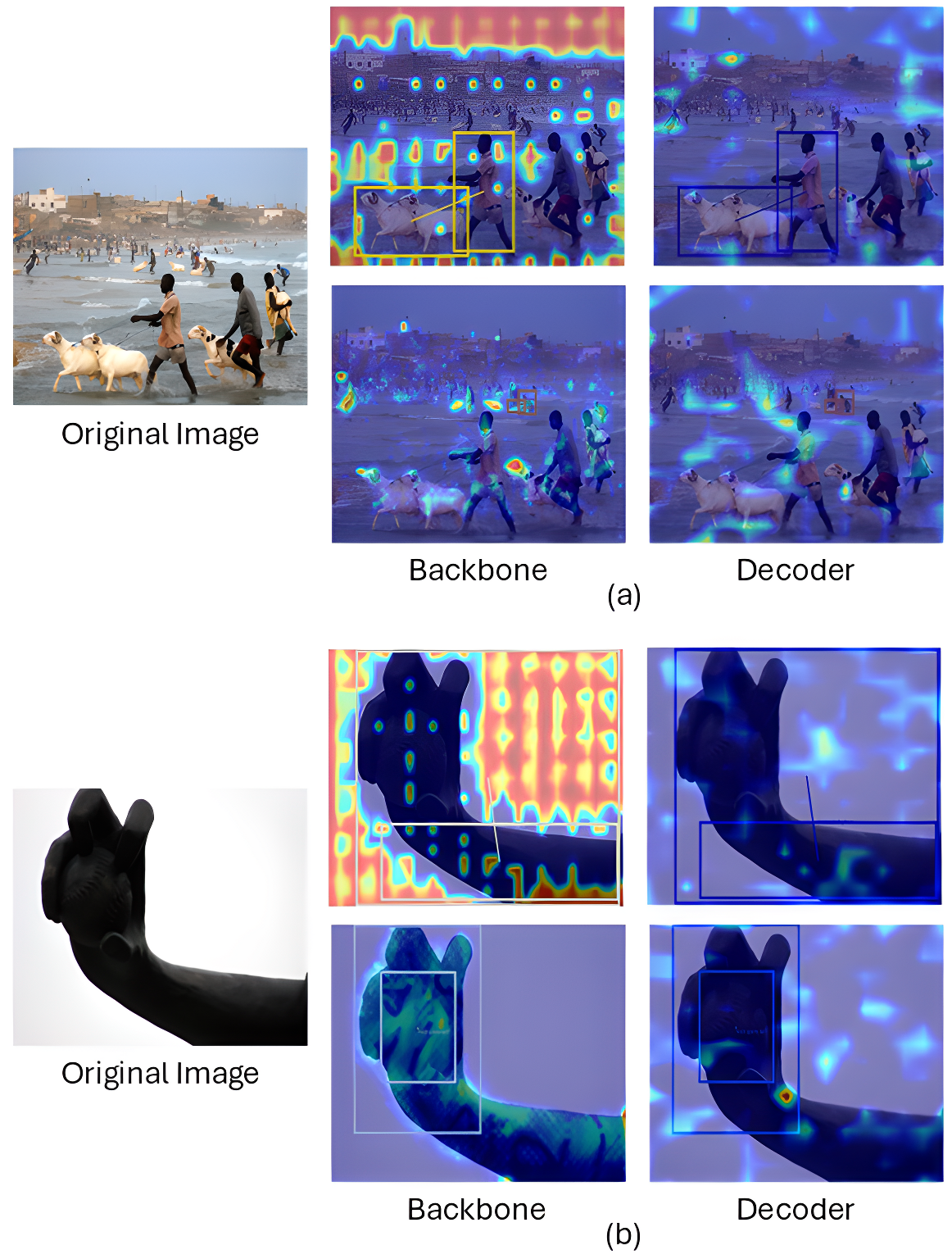}
    \caption{Visualization of Eigen-CAM activation maps on challenging test images from the HICO-DET dataset. The top row represents the heatmap generated from FGAHOI, while the bottom row illustrates the heatmap generated using the proposed architecture.}
    \label{fig:hoi_full}
\end{figure}

To further validate the efficacy of our architecture, we compare it with the FGAHOI model~\cite{fgahoi} using two challenging images, as shown in Fig.~\ref{fig:hoi_full}. The misdirected focus of the Swin backbone results in incorrect predictions. Specifically, it incorrectly identifies ``wash sheep" as ``run dog" in Fig.~\ref{fig:hoi_full}(a), and ``hold sport ball" as ``hold baseball glove" in Fig.~\ref{fig:hoi_full}(b). These misclassifications underscore the critical need for a backbone that can accurately discern and prioritize relevant semantic attributes for precise human-object interaction (HOI) predictions. Our proposed model adeptly localizes key features within images, thereby consistently delivering accurate predictions even in challenging scenarios.

\section{Conclusion and Future Directions}\label{sec:conclusion}
This paper conceptualizes a novel approach for human-object interaction detection, leveraging multi-scale wavelet attention and ray-based encoding. At the heart of our approach lies the wavelet convolution operation, designed to extract middle-order interaction features by aggregating contextual information across diverse filter sizes. This is followed by a novel ray-based encoding designed to compute regions of interest through attention based on ray properties. When incorporated into image classification and HOI detection, the proposed wavelet and ray-based encoding approach efficiently harnesses discriminative middle-order interactions and emphasizes important features. However, the proposed ray-based mechanism is sensitive to training data. As a whole, the ability of multi-scale wavelet combined with a ray-based encoding mechanism has the potential to yield beneficial impacts for image classification and HOI detection. For future research, a number of further improvements can be attained to solidify the overall framework.
\begin{enumerate}
    \item \textbf{Ray Parameterization and Directionality}: Currently, the ray is anchored at the origin but lacks a specified direction, causing it to diffuse across the entire 2D coordinate system instead of maintaining a focused path. As a result, the ray-based layer is sensitive to the data as illustrated in Fig.~\ref{fig:ray_convergence}. To enhance its versatility, the ray can be parameterized with an angle or as a normalized direction vector and positioned based on the learned relationship between the pixels. For instance, employing polar coordinates or direction cosines could enable more flexible and precise control over the ray's trajectory, thereby improving its adaptability to different scenarios.
    \item \textbf{Exploration of Discriminative Features}: The current approach hard-codes the ray onto the 2D coordinate system with the origin as a learnable parameter alongside the pixel coordinates, severely limiting the model's ability to explore the positions of discriminative features. To overcome this limitation, a sampling mechanism within normalized coordinates could be implemented. This mechanism would sample points along the ray in conjunction with pixel interactions, regardless of the image resolution, facilitating a more comprehensive exploration of feature space and leading to improved representation learning.
    \item \textbf{Guidance in Ray Learning}: The current method of learning the ray solely through backpropagation using cross-entropy loss offers no guidance on optimal ray placement within the 2D coordinate system. To enhance learning efficiency and accuracy, additional guidance should be integrated. This can be achieved by reformulating the loss function to incorporate positional guidance or employing a flow network as a prior distribution for the ray position. This approach would enable the model to learn a representative posterior distribution that accurately reflects the optimal positioning of the ray, thus improving overall performance.
\end{enumerate}
By implementing these refinements, the proposed framework can significantly enhance rays' adaptability and representation learning capabilities, leading to more robust performance in various applications.
\section*{Acknowledgment}

This work was supported in part by the Advanced Computing Platform at Monash University Malaysia. The authors would like to thank the anonymous reviewers for their constructive comments and feedback.

\bibliography{section/ref}

\end{document}